\documentclass{fairmeta}

\newif\ifwithappendix
\ifdefined\disableappendix
  \withappendixfalse
\else
  \withappendixtrue
\fi

\usepackage[utf8]{inputenc} 
\usepackage{amsmath}
\usepackage{amssymb}
\usepackage{amsfonts}       
\usepackage{wrapfig}        
\usepackage{float}          

\title{USS: Unifying Spatial-Semantic Prompting for End to End Embodied Visual Tracking}

\author[1,*]{Yuchen Xie}
\author[1,*]{Xinyu Zhou}
\author[1]{Kuangji Zuo}
\author[1]{Yanshuo Lu}
\author[1]{Fengrui Huang}
\author[1]{Boyu Ma}
\author[1,\dagger]{Jianfei Yang}

\affiliation[1]{Nanyang Technological University}

\contribution[*]{Equal Contribution}
\contribution[\dagger]{Corresponding Author}

\correspondence{Jianfei Yang at \email{jianfei.yang@ntu.edu.sg}}

\abstract{
Embodied Visual Tracking (EVT) requires an agent to continuously follow a designated target while moving through dynamic environments. Existing embodied tracking methods generally rely on either an implicit target-selection convention or a language description, leaving the target-specification interface largely fixed. However, different tracking scenarios naturally favor different forms of target specification: language can specify a target outside the robot's current view, whereas spatial prompts provide direct instance designation for visible targets and can be useful for selecting among similar-looking people, designating hard-to-describe individuals, or specifying a target under time pressure. We therefore introduce \textbf{unified spatial-semantic prompting} for EVT, in which text, a point, a box, and a mask serve as complementary target specifications that can be selected according to different scenarios, and present \textbf{USS}, an end-to-end architecture that maps any of them to egocentric waypoints. A modality-specific prompt encoder feeds a common design comprising hybrid-attention fusion, temporal memory, cross-view aggregation, latent prediction, and waypoint decoding, with one policy instance trained for each interface under an identical recipe. Across 320 zero-shot real-robot trials with policies trained only in simulation, spatial prompts perform comparably to language in ordinary tracking scenarios while providing clear benefits when visually similar targets require precise instance designation, supporting our premise that different scenarios favor different target specifications. On simulation EVT-Bench under the standard language-prompt protocol, USS obtains the highest success rate among non-MLLM methods at 57 FPS, against the 4.8--10 FPS reported by MLLM trackers that are stronger on several metrics. Project site: \url{https://arescheah.github.io/uss-project-page/}.
}

\begin{document}

\maketitle

\section{Introduction}
\label{sec:intro}
\begin{figure}[t]
  \centering
  \includegraphics[width=\linewidth,trim=6 10 4 4,clip]{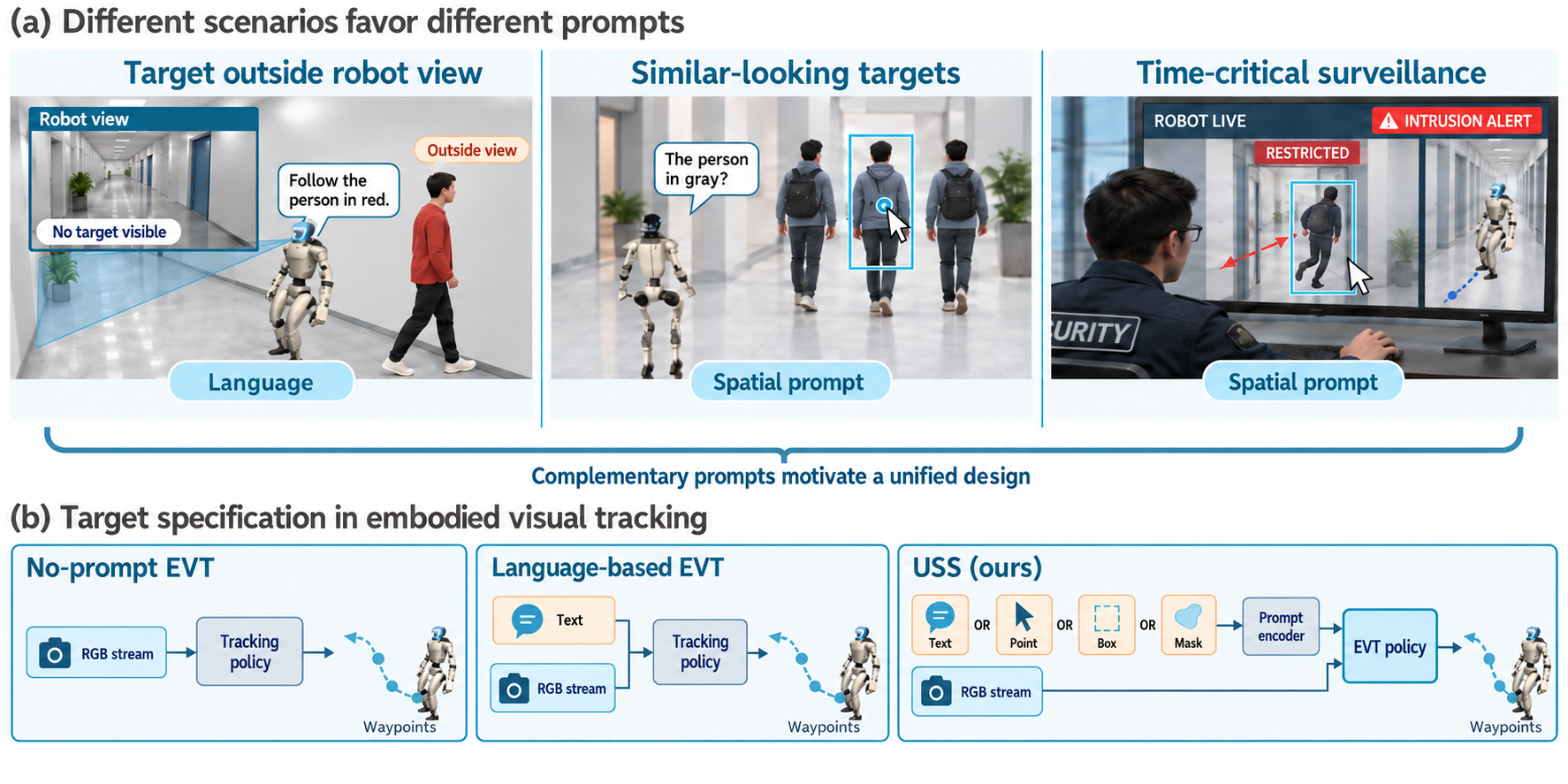}
  \caption{\textbf{Motivations of USS.} (a) Different scenarios favor different specifications: a description can be ambiguous, unavailable, or too slow, while a spatial prompt requires the target to be already visible. Illustrative elements were generated with ChatGPT Image 2.0 and then edited by the authors. (b) Prior EVT policies take no prompt or a language prompt. USS encodes a spatial prompt and conditions the policy on it directly.}
  \label{fig:motivation}
\end{figure}

Embodied Visual Tracking (EVT) aims to enable an embodied agent to follow a specified target in dynamic environments through visual observations~\citep{zhong2024empowering,wang2025trackvla,luo2019end,zhong2019ad,zhang2018coarse,liu2024kurl,zhong2023rspt}. It is a fundamental capability for mobile robots, guide assistants, and other embodied systems, and is inherently closed-loop, since the agent's actions affect its future observations.

Prior EVT methods have progressed from classical visual servoing~\citep{chaumette2006visual}, through end-to-end reinforcement learning~\citep{luo2019end,zhong2019ad,zhang2018coarse} and modular systems that place visual foundation models before downstream control~\citep{zhong2024empowering,liu2024kurl,zhong2023rspt,zeng2024poliformer}, to MLLM-based trackers trained with large-scale imitation learning~\citep{wang2025trackvla,zhang2025embodied,zhang2024uni,zhang2025trackvlapp}, following the broader vision-language-action paradigm for manipulation and generalist control~\citep{black2025pi_,kim2024openvla,black2024pi_0}. While these studies have substantially improved how an embodied agent follows a target, comparatively little attention has been paid to a more fundamental interface question: how should the target itself be specified?

Existing EVT systems are generally tied to a fixed target-specification interface, using either an implicit selection convention or a language description. However, different tracking scenarios naturally favor different forms of target specification. As illustrated in Figure~\ref{fig:motivation}(a), language can specify a target even when it is outside the robot's current view, whereas a spatial prompt can directly designate a visible physical instance. This direct designation is particularly useful when visually similar candidates are difficult to distinguish through description, when the target is hard to describe concisely, or when rapid selection is required in a time-critical situation. Spatial prompting, however, requires the target to be visible at initialization. Language and spatial prompts therefore provide complementary capabilities rather than competing alternatives: no single interface is naturally best suited to every EVT scenario.

We therefore introduce \textbf{unified spatial-semantic prompting} as a new target-specification formulation for EVT. A person designates the target once at initialization using text, a point, a bounding box, or a mask, with the prompt modality selected according to the scenario. The chosen specification is encoded once and conditions the tracking policy throughout the episode. Promptable vision, grounding, and tracking models have already shown that points, boxes, and masks provide effective object-level priors~\citep{kirillov2023segment,zou2023segment,ravi2025sam,li2022grounded,yang2023track,yang2023fine,chen2025improving,zhu2023visual}. Our goal is to bring these complementary interfaces into closed-loop embodied tracking within a common formulation, where an initial grounding error affects every subsequent closed-loop decision.

To realize this formulation, we present \textbf{USS}, an end-to-end architecture that maps any supported target specification directly to egocentric waypoints. A modality-specific prompt encoder converts text, point, box, or mask inputs into target-conditioned representations, which interact with visual observations through hybrid-attention fusion. Temporal memory preserves target information over time, cross-view aggregation combines complementary camera observations, and learnable queries distill the fused representation into a sparse state for waypoint prediction. USS thus provides a common architecture for all four prompting interfaces, with one policy instance trained for each modality under an identical recipe. This common design keeps the downstream tracking objective unchanged across all prompt modalities.

USS further incorporates an action-conditioned latent predictor~\citep{assran2023self,bardes2024revisiting,assran2025v,mur2026v} as an auxiliary training objective. It conditions the current sparse state on the predicted waypoint sequence and estimates the next latent state, aligning the prediction with a detached target from an exponential-moving-average encoder rather than reconstructing pixels. The predictor is discarded at inference.

Our contributions are summarized as follows.
\begin{itemize}
    \item \textbf{Unified spatial-semantic prompting for EVT.}
    We formulate target specification as a scenario-dependent choice among complementary interfaces: language, point, bounding box, and mask. Rather than assuming one fixed interface, the prompt modality can be selected according to the demands of the tracking scenario.

    \item \textbf{USS, an end-to-end architecture supporting semantic and spatial prompting.}
    USS maps text, point, box, or mask specifications directly to egocentric waypoints through modality-specific prompt encoding, hybrid-attention fusion, temporal memory, cross-view aggregation, and sparse-query waypoint prediction, with an auxiliary action-conditioned latent predictor used only during training.

    \item \textbf{Validation on a physical robot and on EVT-Bench.}
    Across 320 zero-shot real-robot trials with policies trained only in simulation, spatial prompts perform comparably to language in ordinary tracking scenarios while providing clear benefits when visually similar targets require precise instance designation. These results support our premise that different scenarios favor different target specifications. Under the standard language-prompt protocol on EVT-Bench, USS achieves the highest success rate among non-MLLM methods while running at 57 FPS.
\end{itemize}

\section{Related Work}
\noindent\textbf{Human-Robot Interaction.} Although natural language has dominated embodied AI interfaces, deictic and multimodal cues can identify a referent without a complete verbal description~\citep{saran2018human,bar2022visual,campagna2025fostering,green2025using,park2024towards,yang2023set}. For direct human-in-the-loop control, NMM-HRI, FAM-HRI, and GesVLA combine speech with forearm posture, gaze, or pointing~\citep{lai2025natural,lai2026fam,guo2026gesvla,choi2024gaze}, while in autonomous policy design VP-VLA, MOKA, and TraceVLA use visual marks, coordinate annotations, or motion traces as interfaces for policy execution~\citep{wang2026vp,liu2024moka,zheng2025tracevla,nasiriany2024pivot}. Both lines have been studied mainly in tabletop manipulation and generalist vision-language-action (VLA) control; USS brings such designation to EVT, where the designated instance must be held under ego-motion and followed over time.

\noindent\textbf{Embodied Visual Tracking.} Embodied visual tracking couples target recognition with closed-loop motion control from egocentric observations. Early methods relied on end-to-end RL, sometimes with adversarial multi-agent training~\citep{luo2019end,zhong2019ad,zhang2018coarse}, but the learned visual representations were often insufficient for robust target identity preservation. Later modular systems use a visual foundation model to turn a target description into masks and pass those masks, recomputed every frame, to a downstream planner~\citep{zhong2024empowering,liu2024kurl,zhong2023rspt}. There a box or a point is, like a sentence, an argument to the segmenter, whose per-frame output is all the planner receives, and errors crossing that perception-control interface accumulate over time. Recent VLA trackers~\citep{wang2025trackvla,zhang2025embodied,zhang2024uni,zhang2025trackvlapp} use large-scale imitation learning on MLLM backbones to reduce the perception-control interface mismatch, at the cost of lower inference throughput. USS instead conditions a single end-to-end policy on language, box, point, or mask prompts, so the designation never has to cross a per-frame perception-control interface, while keeping real-time throughput.
\begin{figure}[ht]
    \centering
    \includegraphics[width=\linewidth]{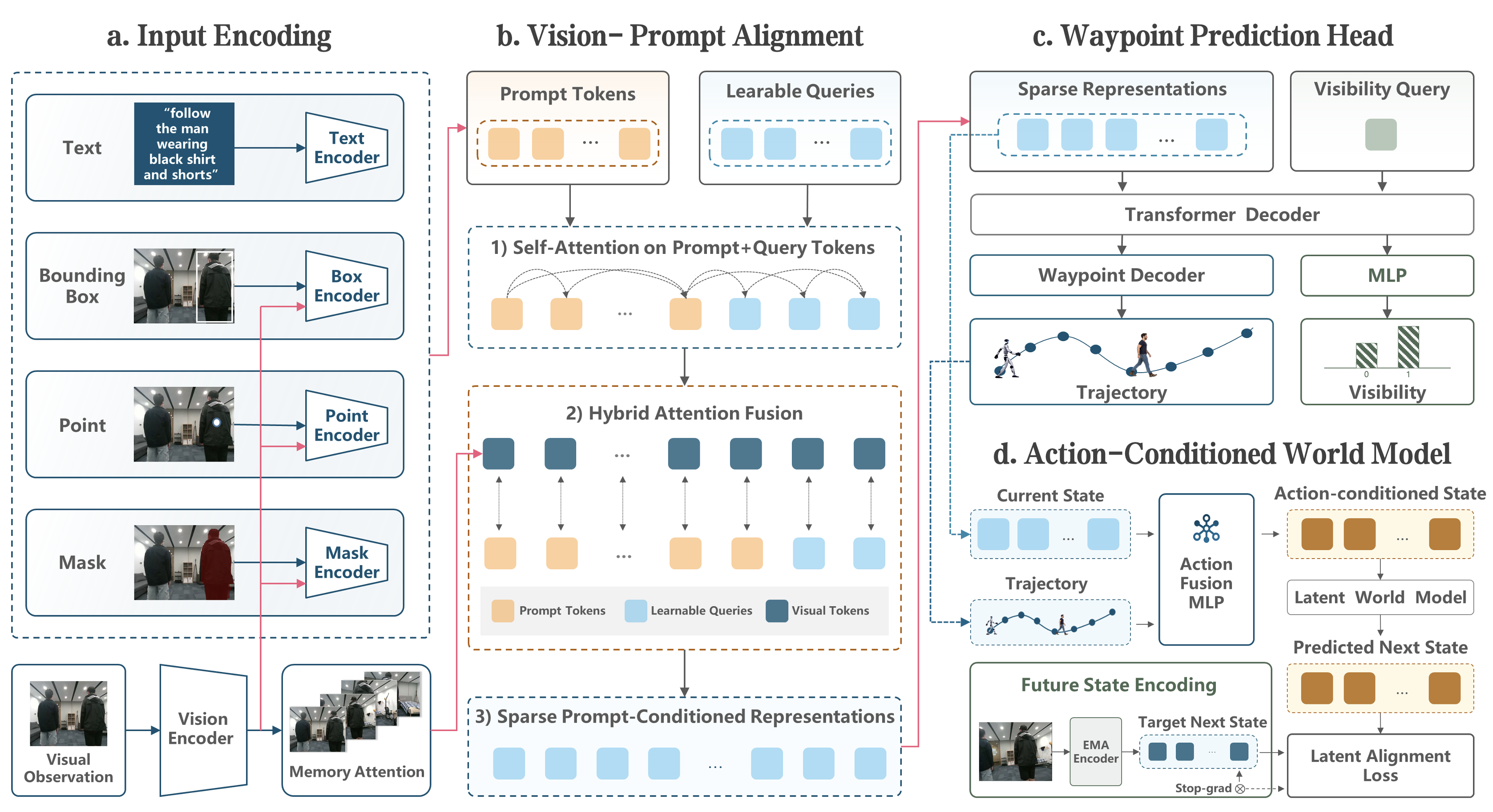}
\caption{The common USS architecture, which maps RGB observations and a target prompt to egocentric waypoints. During training, an action-conditioned latent world model further regularizes future state prediction to improve temporal consistency.}
    \label{fig:method}
\end{figure}

\section{Method}
USS maps an RGB observation window and one target specification to egocentric waypoints. It is one architecture, shown in Figure~\ref{fig:method}, instantiated once per prompt modality, and it consists of a vision-prompt encoder with temporal memory, a waypoint decoder, and an auxiliary action-conditioned latent predictor used only during training.

\subsection{Problem Formulation}
At time $t$, the robot receives a synchronized multi-view RGB image window $\mathcal{I}_{t-S+1:t}=\{I_\tau^{(n)}\mid \tau=t-S+1,\ldots,t;\ n=1,\ldots,N\}$ and a raw target prompt $\mathcal{G}$, where $S$ is the temporal window length, $N$ is the number of camera views, and $I_\tau^{(n)}$ denotes the image from camera $n$ at time $\tau$. The prompt $\mathcal{G}$, which is language, a box, a point, or a mask, is supplied once, at $t{=}1$, by the person operating the robot; it designates which instance the episode is about, not where that instance lies at later steps. It is encoded once into a target specification $Y^{\gamma}$, where $\gamma$ denotes the prompt modality, and held fixed thereafter. Let $\theta_\gamma$ denote the weights of the USS instance trained for modality $\gamma$. It predicts $M$ future egocentric waypoints $\hat{W}_t=\pi_{\theta_\gamma}(\mathcal{I}_{t-S+1:t},Y^{\gamma},\mathcal{P}_t)$, where $W_t=[w_t,\dots,w_{t+M-1}]$, $w_t=(x_t,y_t)\in\mathbb{R}^2$, and optional multi-view camera poses $\mathcal{P}_t\in\mathbb{R}^{N\times7}$ provide 7D pose calibration for the $N$ views. The objective is to hold the designated identity and maintain a following distance of $d\in[1,3]$ meters.

\subsection{Vision-Prompt Encoder}

\paragraph{Visual Encoder with Temporal Memory.} The visual encoder processes each camera stream independently before cross-view aggregation. For image $I_t^{(n)}$, a pretrained PE-Spatial encoder~\citep{bolya2026perception} produces dense patch tokens $V_t^{(n)}=\mathcal{E}_{\mathrm{vis}}(I_t^{(n)})\in\mathbb{R}^{P\times C_v}$, where $P$ is the number of image patches and $C_v$ is the visual feature dimension. All but the last two encoder blocks are frozen.

To provide short-term temporal context, each view keeps a sliding memory bank $\mathcal{M}_t^{(n)}$ of recent visual tokens. We add a sinusoidal time encoding~\citep{vaswani2017attention}, concatenate the encoded memory tokens as $K_t^{(n)}$, and let the current tokens attend to them:
\begin{equation}
Z_t^{(n)}=\mathrm{FFN}\!\left(\mathrm{CrossAttn}(\mathrm{SelfAttn}(V_t^{(n)}),K_t^{(n)})\right).
\end{equation}
A mask prompt additionally stores the first-frame mask-enhanced feature from the prompted view as a persistent target anchor in this memory.

\paragraph{Prompt Encoder.}
The prompt encoder maps the person's designation, supplied once at initialization, into a representation compatible with the visual feature space. We use $Y^{\gamma}$ to denote the representation produced for prompt type $\gamma\in\{\mathrm{text},\mathrm{box},\mathrm{point},\mathrm{mask}\}$. For language prompts, let $s_{1:L}$ denote the $L$ input word tokens. A frozen PE-Core text encoder~\citep{bolya2026perception} $\mathcal{E}_{\mathrm{text}}$ outputs text-token embeddings $h_{\mathrm{cls}},h_1,\ldots,h_L$, where $h_{\mathrm{cls}}$ is the global \texttt{[CLS]} embedding and $h_i$ is the embedding of word token $s_i$. We retain the \texttt{[CLS]} token and nominal tokens indexed by $\mathcal{J}_{\mathrm{nom}}$ as text prompt tokens:
\begin{equation}
    Y^{\mathrm{text}}=[h_{\mathrm{cls}};\{h_i\}_{i\in\mathcal{J}_{\mathrm{nom}}}],\quad
    [h_{\mathrm{cls}},h_1,\ldots,h_L]=\mathcal{E}_{\mathrm{text}}(s_{1:L}).
\end{equation}
The person places a box or point on the prompted view $n$ at $t{=}1$, so the encoder reads that frame's visual tokens $V_1^{(n)}$. Let $R^g$ denote the designated region for prompt type $g\in\{\text{box},\text{point}\}$; for a point prompt, $R^g$ is a fixed-size pseudo box centered at the designated pixel. Both are encoded by applying RoIAlign~\citep{he2017mask} over the spatial token grid of $V_1^{(n)}$, followed by a prompt-specific pooling/projection head $\psi_g$ and a learnable grid embedding $E_g$:
\begin{equation}
    Y^g=\psi_g(\mathrm{RoIAlign}(V_1^{(n)},R^g))+E_g,\qquad g\in\{\mathrm{box},\mathrm{point}\}.
\end{equation}
For mask prompts, let $m_1$ denote the binary mask the person supplies in view $n$ at $t{=}1$. A lightweight mask encoder $\phi_{\mathrm{mask}}$ converts $m_1$ into a dense spatial prior $B^{\text{mask}}$, which is flattened and added to the first-frame visual tokens $V_1^{(n)}$:
\begin{equation}
    B^{\text{mask}}=\phi_{\mathrm{mask}}(m_1),\qquad
    Y^{\mathrm{mask}}=V_1^{(n)}+\mathrm{Flatten}(B^{\text{mask}}).
\end{equation}
Every $Y^{\gamma}$ is computed once, at initialization, and reused unchanged, with $Y^{\mathrm{text}}$, $Y^{\mathrm{box}}$, and $Y^{\mathrm{point}}$ used as prompt-token sequences that an MLP projects into the visual embedding space before fusion, and $Y^{\mathrm{mask}}$ used as the dense mask-enhanced anchor stored in memory. Thereafter only the RGB stream is re-encoded.

\paragraph{Vision-Prompt Fusion.}
The fusion encoder is where prompt information and visual evidence interact. For each view, let $Y^{(n)}$ denote the projected prompt-token sequence, the same at every step; for mask prompts the prompt enters through the memory anchor $Y^{\mathrm{mask}}$. We introduce $K_q$ learnable abstraction queries $q^{(n)}\in\mathbb{R}^{K_q\times C_v}$, which distill a compact target-conditioned representation for control, and concatenate them with the prompt tokens as $u^{(n)}=[q^{(n)};Y^{(n)}]$. Self-attention over these tokens lets prompt evidence and queries exchange information before reading from the image:
\begin{equation}
    u^{\prime(n)}=\mathrm{SelfAttn}\!\left(u^{(n)}\right).
\end{equation}
The fusion encoder then performs bidirectional attention between the compact prompt/query tokens and the dense visual tokens. It follows a read-write-read pattern in which the prompt/query tokens first read relevant evidence from $Z_t^{(n)}$, the updated tokens then write prompt-conditioned information back to the visual stream to emphasize designated regions and suppress distractors, and finally the queries read the updated visual representation to produce sparse tokens:
\begin{align}
    u_t^{r(n)} &= \mathrm{FFN}\!\left(\mathrm{CrossAttn}_{u \leftarrow Z}\!\left(u^{\prime(n)},\, Z_t^{(n)}\right)\right), \\
    Z_t^{f(n)} &= \mathrm{CrossAttn}_{Z \leftarrow u}\!\left(Z_t^{(n)},\, u_t^{r(n)}\right), \\
    \hat{u}_t^{(n)} &= \mathrm{CrossAttn}_{u \leftarrow Z}\!\left(u_t^{r(n)},\, Z_t^{f(n)}\right).
\end{align}
Here $u_t^{r(n)}$ and $\hat{u}_t^{(n)}$ are intermediate prompt/query tokens, and $Z_t^{f(n)}$ is the fused dense visual token sequence; the time index enters only through $Z_t^{(n)}$. We retain the first $K_q$ query outputs, $Q_t^{(n)}=\hat{u}_t^{(n)}[1{:}K_q]$, as the sparse prompt-conditioned representation passed to the waypoint head.

\subsection{Waypoint Prediction Head}
Before decoding waypoints, we optionally add a 3D positional encoding to the fused visual tokens. Following PETR~\citep{liu2022petr}, we lift image tokens into a camera-aware 3D reference space using calibration information and encode the resulting geometry with a small MLP, giving tokens from different cameras a shared robot-frame spatial prior. When camera poses are unavailable, we simply use the original fused visual tokens. We denote the resulting dense token sequence as $\tilde{Z}_t^{(n)}$. A transformer decoder shared across camera views then maps the sparse query tokens $Q_t^{(n)}$ and a learnable presence query $q_{\mathrm{pres}}^{(n)}$ to a view-level state $\hat{S}_t^{(n)}$ and a view-wise visibility prediction $\hat{v}_t^{(n)}$:
\begin{equation}
  [\hat{S}_t^{(n)};\hat{v}_t^{(n)}]=\mathrm{Dec}([Q_t^{(n)};q_{\mathrm{pres}}^{(n)}],\tilde{Z}_t^{(n)}).
\end{equation}
All view-level states are concatenated into $S_t\in\mathbb{R}^{N K_q\times C_v}$, from which a waypoint decoder predicts the future trajectory $\{\hat{w}_{t+i}\}_{i=0}^{M-1}$. With ground-truth waypoints $w^*_{t+i}$, binary view-wise visibility labels $v_t^{*(n)}$, and binary cross entropy $\mathrm{BCE}(\cdot)$, the supervised losses are:
\begin{equation}
    \mathcal{L}_{\mathrm{wp}}=\frac{1}{2M}\sum_{i=0}^{M-1}\|\hat{w}_{t+i}-w^*_{t+i}\|_1,\qquad
    \mathcal{L}_{\mathrm{pres}}=\mathrm{BCE}(\{\hat{v}_t^{(n)}\}_{n=1}^{N},\{v_t^{*(n)}\}_{n=1}^{N}).
\end{equation}

\subsection{Action-Conditioned World Model}
Following V-JEPA 2-AC~\citep{assran2025v}, we train an auxiliary action-conditioned latent world model that predicts the next sparse state rather than reconstructing pixels. We flatten predicted waypoints into an action vector $a_t=\mathrm{vec}(\hat{W}_t)$, repeat it across the $N K_q$ sparse tokens, and use an action-fusion MLP $\mathrm{MLP}_a$ to produce the action-conditioned state $G_t$:
\begin{equation}
  G_t=\mathrm{MLP}_a(\mathrm{Concat}(S_t,\mathrm{Repeat}(a_t,N K_q))),\qquad
  \widetilde{S}_{t+1}=\mathcal{F}_\phi(G_t).
\end{equation}
The predictor $\mathcal{F}_\phi$ maps $G_t$ to the next sparse state prediction $\widetilde{S}_{t+1}$. To stabilize targets, we maintain an EMA copy of the representation pathway with parameters $\theta_{\mathrm{ema}}$, updated from online parameters $\theta$ using momentum $\mu$ as $\theta_{\mathrm{ema}}\leftarrow \mu\theta_{\mathrm{ema}}+(1-\mu)\theta$. We denote this EMA representation encoder by $E_{\theta_{\mathrm{ema}}}$, and use its detached next-step sparse state $S^{\mathrm{ema}}_{t+1}=E_{\theta_{\mathrm{ema}}}(\{I_{t+1}^{(n)}\}_{n=1}^{N},Y^{\gamma})$ as the target. The EMA branch runs the same prompt encoder, memory, fusion, and decoder as the online pathway, so prediction and target lie in the same prompt-conditioned space. Let $\bar{S}_{t+1}=\mathrm{LN}(\widetilde{S}_{t+1})$ and $\bar{S}^{\mathrm{ema}}_{t+1}=\mathrm{sg}(\mathrm{LN}(S^{\mathrm{ema}}_{t+1}))$ denote the normalized prediction and stop-gradient target. The latent prediction loss is a normalized SmoothL1 alignment:
\begin{equation}
  \mathcal{L}_{\mathrm{wm}}=\frac{1}{N K_q C_v}\sum_{i=1}^{N K_q}\sum_{c=1}^{C_v}
  \rho_\beta\!\left(\bar{S}_{t+1,i,c}-\bar{S}^{\mathrm{ema}}_{t+1,i,c}\right),
\end{equation}
where $\mathrm{LN}$ is per-token layer normalization, $\mathrm{sg}$ is stop-gradient, and $\rho_\beta$ is SmoothL1 with transition parameter $\beta$.

\subsection{Overall Training Objective}
Each policy instance $\pi_{\theta_\gamma}$ is trained end-to-end with trajectory imitation, view-wise visibility prediction, and latent dynamics modeling. The scalar weights $\lambda_{\mathrm{pres}}$ and $\lambda_{\mathrm{wm}}$ balance the visibility and world-model losses:
\begin{equation}
  \mathcal{L}=\mathcal{L}_{\mathrm{wp}}+\lambda_{\mathrm{pres}}\mathcal{L}_{\mathrm{pres}}+\lambda_{\mathrm{wm}}\mathcal{L}_{\mathrm{wm}}.
\end{equation}
The EMA pathway and world-model predictor are used only during training; inference uses only prompt-conditioned perception and the waypoint head, so the auxiliary branch adds no runtime overhead.

\section{Experimental Results}
\label{sec:result}
We evaluate USS along three axes. We first deploy it on a physical robot to examine prompt flexibility and closed-loop robustness under real sensing and control noise. We then evaluate standardized simulation performance on EVT-Bench to compare against existing EVT systems under the same protocol. Finally, we conduct controlled simulation ablations to isolate the effects of cross-view fusion, temporal memory, latent dynamics learning, and action decoding.

\subsection{Experimental Setup}
\label{sec:setup}
\paragraph{View configuration.}
USS is designed for multi-view perception, with per-view encoding and fusion, a PETR-style 3D positional encoding that places all cameras in one robot-frame reference space, and a per-view presence head. On EVT-Bench we therefore run USS with the three egocentric cameras from which the training data is rendered, on the benchmark's standard episodes and metrics; the two strongest MLLM trackers, NavFoM and TrackVLA++, likewise report their best results with four views. Extra cameras are not a free gain, since a policy can use them only if its architecture aggregates across views: four views over one improves NavFoM's DT success rate by 0.6 points and TrackVLA++'s by 7.5. The modular baselines are single-view, so the additional views enter that comparison as part of the architecture USS contributes rather than as a change to the protocol; Table~\ref{tab:ablation_dt_bbox} puts cross-view fusion at 5.5 SR on DT, a fraction of USS's 27.9--34.1-point lead over those baselines, so the lead reflects the architecture as a whole rather than the view count. The real robot carries one camera, so all real-robot results use a single-view instance of the same architecture, recipe, and supervision.

\paragraph{Training data and supervision.}
We construct the simulation training set from the training branch of EVT-Bench~\citep{wang2025trackvla}, inheriting its humanoid assets, target identities, and route specifications. We select the 800 episodes with the longest target routes and replay them in Habitat using three egocentric cameras rendered at 10 Hz and $512\times512$ resolution. At each frame, a teacher tracker picks a following point behind the target, rolls out a short collision-aware path on the navigation mesh, and transforms the planned positions into the current tracker frame to supervise future waypoints. View-wise visibility labels are derived from seven approximate humanoid keypoints, a keypoint counting as visible only inside the camera frustum and after a ray-based occlusion test.\ifwithappendix\ Appendix~\ref{app:data_collection} gives the full data-collection procedure.\fi

\paragraph{Prompt generation and optimization.}
Language instructions are taken from the original benchmark annotations. For spatial-prompt training, we adjust the initial heading so that the target is visible but not perfectly centered. Bounding boxes are obtained by projecting an upright 3D cuboid around the target, points by projecting a conservative torso location, and masks by applying SAM2 to the first-frame target region. Each prompt is generated once at initialization and never recomputed. We train one instance of USS per prompt modality under an identical optimization recipe. All use a pretrained PE-Spatial-B16-512 image encoder; the language policy additionally uses a frozen pretrained PE-Core-B16-224 text encoder. The last two image-transformer blocks are fine-tuned and the remainder of the encoder is frozen. We use $K_q=10$ abstraction queries per view, a planner hidden dimension of $C_v=256$, a 16-frame memory bank, and $M=10$ predicted waypoints; the prompt-fusion, memory, and world-model blocks each use 2 layers with 8 heads, and the 3D positional encoding uses 64 depth bins. A box prompt is pooled with $7\times7$ RoIAlign into 9 prompt tokens, and a point prompt into 4 tokens from a $160$\,px pseudo box with $5\times5$ RoIAlign. The loss weights are $\lambda_{\mathrm{pres}}=0.05$ and $\lambda_{\mathrm{wm}}=0.2$. Training runs for 12 epochs with AdamW under its default momentum parameters, a learning rate of $5\times10^{-5}$ with a $0.1$ multiplier on backbone parameters, weight decay $0.01$, cosine decay, batch size 1 per GPU, and gradient-norm clipping at 35. A subsequent DAgger-style stage relabels model-driven rollouts with the same teacher to reduce imitation-learning distribution shift.\ifwithappendix\ Appendix~\ref{app:training_details} lists the complete hyperparameters.\fi

\paragraph{Closed-loop inference.}
USS predicts a robot-frame waypoint sequence and view-wise target-presence logits. At each control step, only the first waypoint is executed. In simulation it is transformed to world coordinates and snapped to the navigation mesh. If every view has presence probability below $0.5$, the controller may reuse the next waypoint from the preceding prediction, bridging short disappearances without re-designating the target. On hardware the policy is stepped at its measured inference rate. The EMA encoder and latent predictor are removed at inference time.

\subsection{Real-World Experiments}
\label{sec:real}
\begin{figure}[t]
  \centering
  \includegraphics[width=\linewidth]{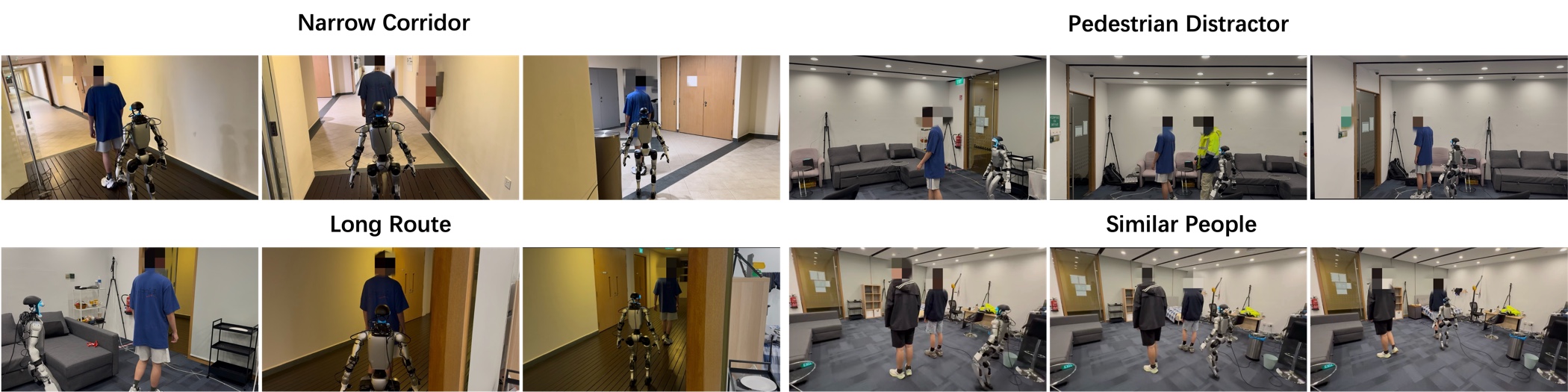}
  \caption{Real-world tracking rollouts across four indoor scenes. Each scene is shown as a $1\times3$ frame sequence, ordered chronologically from left to right. Egocentric views with the prompt overlaid, and complete rollouts, are provided on the project site.}
  \label{fig:real_world_qualitative}
\end{figure}

\begin{wraptable}{r}{0.44\textwidth}
\vspace{-1.0em}
\centering
\caption{Real-world success rate (SR). Each scene is evaluated over 20 trials.}
\label{tab:real_world_quant}
\fontsize{8.5pt}{10pt}\selectfont
\setlength{\tabcolsep}{1.8pt}
\renewcommand{\arraystretch}{1.05}
\begin{tabular}{@{}lcccc@{}}
\toprule
Scene & Text & BBox & Point & Mask \\
\midrule
Narrow corridor & \textbf{100\%} & \textbf{100\%} & \textbf{100\%} & \textbf{100\%} \\
Ped. distractor & 75\% & \textbf{85\%} & \textbf{85\%} & 80\% \\
Long route & 70\% & \textbf{85\%} & 80\% & 80\% \\
Similar people & 45\% & \textbf{90\%} & 80\% & 70\% \\
\bottomrule
\end{tabular}
\vspace{-1.2em}
\end{wraptable}
We deploy USS on a Unitree G1 equipped with a chest-mounted Intel RealSense D455 camera. RGB observations are transmitted over Ethernet to an RTX 4090 server, which runs USS and returns high-level velocity commands to the robot-side gait controller. Planar velocity comes from the first waypoint displacement and yaw rate from the final waypoint heading, so USS handles target-conditioned perception and following decisions while the onboard controller handles locomotion.\ifwithappendix\ Appendix~\ref{app:real_world_deployment} details the platform and protocol.\fi

We evaluate narrow-space following, obstacle-rich navigation with unrelated pedestrians, longer routes with turns, and a semantically ambiguous setting where multiple people have similar appearance or language descriptions. For each scene and prompt type, we conduct 20 trials, 320 in total. All real-robot results use the simulation-trained single-view policy zero-shot, with no real-world data, fine-tuning, or calibration. One prompt is supplied at the start of the trial with no later correction, and in every condition the target is then visible and unoccluded, matching the spatial-prompt initialization of Section~\ref{sec:sim}, so the conditions differ only in how the instance is designated. In the similar-person scene the description identifies the target uniquely, since both candidates wear a black top and only the trousers differ, so a language failure there is a grounding failure. A trial fails if the robot switches to another person, or turns away and then either loses the target for more than five seconds or never recovers the heading needed to resume; all remaining trials count as successes. Figure~\ref{fig:real_world_qualitative} shows representative rollouts, and Table~\ref{tab:real_world_quant} reports success rates.

Across the tested scenes the prompt types behave similarly except where a distractor shares most of the target's described attributes. All four solve the narrow corridor, where the target is isolated and continuously visible. Under pedestrian distractors and on the long route the spatial prompts lead by one to three trials out of twenty, a margin we do not read as a difference between interfaces. The similar-person scene separates them. The bounding-box prompt succeeds in 18 of 20 trials against 9 of 20 for language, with the point and mask prompts at 16 and 14. Because a spatial prompt can only be placed on a visible instance, the spatial conditions begin from a reference that is already resolved while the language condition must resolve it from the description alone. That asymmetry is the property under test, so the margin measures how reliably a sufficient description is grounded and held in this scene rather than a general advantage of spatial prompts.

\subsection{Simulation Benchmark Comparison}
\label{sec:sim}
We evaluate simulation performance of USS on EVT-Bench introduced by TrackVLA~\citep{wang2025trackvla}, which contains three progressively harder tasks: Single-Target Tracking (STT), Distracted Tracking (DT), and Ambiguity Tracking (AT). STT measures basic following, DT adds distractors that test target identity persistence, and AT further stresses ambiguous language specifications. Following the benchmark protocol we report success rate (SR), tracking rate (TR) over timesteps with a valid following relationship, collision rate (CR) over collision-terminated episodes, and inference speed (FPS). Each split provides 1{,}405 held-out test episodes in 101 unseen scenes, and we evaluate every method on the full split. All reported numbers come from a single evaluation run of a single trained instance per prompt modality, so we do not report variance across seeds.

\begin{table}[t]
\centering
\caption{EVT-Bench results (SR/TR/CR; $\uparrow$/$\uparrow$/$\downarrow$). \textcolor{red}{Red} and \textcolor{blue}{blue} mark the best non-MLLM and MLLM language results, and bold marks the best spatial result. $\dagger$: grounding with GroundingDINO; $\ddagger$: with SoM+GPT-4o; \P: four-view setting; \S: hardware given in Section~\ref{sec:sim}.}
\label{tab:evt_bench}
\par\vspace{0.35em}
\fontsize{8.5pt}{10pt}\selectfont
\setlength{\tabcolsep}{2.2pt}
\renewcommand{\arraystretch}{0.93}
\begin{tabular*}{\textwidth}{@{\extracolsep{\fill}}lcccccccccc@{}}
\toprule
\multirow{2}{*}{Methods} &
\multicolumn{3}{c}{\textit{STT}} &
\multicolumn{3}{c}{\textit{DT}} &
\multicolumn{3}{c}{\textit{AT}} &
\multirow{2}{*}{FPS\textsuperscript{\S}$\uparrow$} \\
\cmidrule(lr){2-4}\cmidrule(lr){5-7}\cmidrule(lr){8-10}
 & SR$\uparrow$ & TR$\uparrow$ & CR$\downarrow$ & SR$\uparrow$ & TR$\uparrow$ & CR$\downarrow$ & SR$\uparrow$ & TR$\uparrow$ & CR$\downarrow$ & \\
\midrule
\multicolumn{11}{@{}c@{}}{\textit{Modular language-prompt baselines (non-MLLM)}} \\
\midrule
IBVS$^{\dagger}$~\citep{gupta2016novel} & 42.9 & 56.2 & 3.75 & 10.6 & 28.4 & \textcolor{red}{6.14} & 15.2 & \textcolor{red}{39.5} & \textcolor{red}{4.90} & -- \\
PoliFormer$^{\dagger}$~\citep{zeng2024poliformer} & 4.67 & 15.5 & 40.1 & 2.62 & 13.2 & 44.5 & 3.04 & 15.4 & 41.5 & -- \\
EVT~\citep{zhong2024empowering} & 24.4 & 39.1 & 42.5 & 3.23 & 11.2 & 47.9 & 17.4 & 21.1 & 45.6 & 15.0 \\
EVT$^{\ddagger}$~\citep{zhong2024empowering} & 32.5 & 49.9 & 40.5 & 15.7 & 35.7 & 53.3 & 18.3 & 21.0 & 44.9 & -- \\
\textbf{USS (language)} & \textcolor{red}{70.8} & \textcolor{red}{72.1} & \textcolor{red}{3.13} & \textcolor{red}{49.8} & \textcolor{red}{54.6} & 9.89 & \textcolor{red}{34.2} & 35.8 & 28.2 & \textcolor{red}{57.0} \\
\midrule
\multicolumn{11}{@{}c@{}}{\textit{USS spatial-prompt variants (visible-target initialization; separate protocol)}} \\
\midrule
\textbf{USS (point)} & 83.4 & 86.8 & 3.02 & 79.8 & 80.2 & \textbf{2.92} & -- & -- & -- & 68.0 \\
\textbf{USS (mask)} & 81.3 & 80.8 & 6.65 & 75.8 & 76.3 & 3.04 & -- & -- & -- & \textbf{72.0} \\
\textbf{USS (box)} & \textbf{86.7} & \textbf{92.2} & \textbf{2.73} & \textbf{83.6} & \textbf{81.5} & 2.93 & -- & -- & -- & 65.0 \\
\midrule
\multicolumn{11}{@{}c@{}}{\textit{MLLM-based language-prompt methods}} \\
\midrule
Uni-NaVid~\citep{zhang2024uni} & 25.7 & 39.5 & 41.9 & 11.3 & 27.4 & 43.5 & 8.26 & 28.6 & 43.7 & 5.0 \\
NavFoM\textsuperscript{\P}~\citep{zhang2025embodied} & 88.4 & 80.7 & -- & 62.0 & 67.9 & -- & -- & -- & -- & 5.1 \\
TrackVLA~\citep{wang2025trackvla} & 85.1 & 78.6 & 1.65 & 57.6 & 63.2 & 5.80 & 50.2 & 63.7 & 17.1 & \textcolor{blue}{10.0} \\
TrackVLA++\textsuperscript{\P}~\citep{zhang2025trackvlapp} & \textcolor{blue}{90.9} & \textcolor{blue}{82.7} & \textcolor{blue}{1.50} & \textcolor{blue}{74.0} & \textcolor{blue}{73.7} & \textcolor{blue}{3.51} & \textcolor{blue}{55.9} & \textcolor{blue}{63.8} & \textcolor{blue}{15.1} & 4.8 \\
\bottomrule
\end{tabular*}
\end{table}

Table~\ref{tab:evt_bench} reports three separate comparison blocks. The two language-prompt blocks follow the standard benchmark initialization and distinguish modular non-MLLM pipelines, each pairing a grounding front-end with a downstream controller, from MLLM-based trackers. The spatial-prompt block uses visible-target initialization because an instance can only be designated once it is visible. To avoid an overly favorable spatial setup, we perturb the initial robot heading within $\pm20^\circ$, so the target is visible but not perfectly centered. Because this initialization changes the grounding condition, spatial-prompt results are not used for direct protocol-matched ranking against language-prompt methods, including USS's own. We report them on STT and DT to characterize tracking after an instance has been explicitly selected. We do not report them on AT: in EVT-Bench, AT differs from DT only in that the instruction is replaced by an expression denoting the first person the agent observes, and a spatial prompt designates that instance directly, so an AT column would restate the DT column.

Under the standard language-prompt protocol, USS obtains the highest SR among the modular non-MLLM baselines on STT, DT, and AT, improving on the best prior entry in that group by 27.9, 34.1, and 15.9 points respectively. It also obtains the highest tracking rates on STT and DT, whereas IBVS retains a higher AT tracking rate and lower collision rates on DT and AT. The MLLM block is stronger on success rate across the splits, led by TrackVLA++. We measure 57 FPS for USS on an RTX 4090. TrackVLA, TrackVLA++ and NavFoM report 10, 4.8 and 5.1 FPS on the same GPU, and Uni-NaVid reports 5 Hz on an A100; among the modular baselines, EVT reports 66\,ms per step with DEVA on an RTX 3090, i.e.\ about 15 FPS. NavFoM separately reports at most 0.5\,s per eight-waypoint trajectory when deployed, i.e.\ about two planning steps per second. We do not re-benchmark any of these methods and take their numbers as reported, so we read the gaps as differences in deployment cost rather than controlled latency measurements. At 10 FPS a forward pass occupies an entire 100\,ms control period, and at 57 FPS about a sixth of it. TrackVLA, TrackVLA++ and USS all stream images to an off-board server, so this compares models under a shared deployment pattern rather than on-board against off-board.

Within the separate spatial-prompt protocol the box variant is strongest on both splits, and the point variant obtains the lowest DT collision rate. Even though a mask is the denser and more informative designation, it enters as a dense first-frame anchor in memory and never passes through the prompt-token fusion path that the box and point use, so its density is not matched by greater participation in fusion. Routing the mask through the prompt-token path would separate the architectural factor from the representational one, and we leave this to future work. These numbers characterize USS once an instance has been selected, and as noted above they are not comparable with either language-prompt block. No prior EVT method accepts a spatial prompt, so there is no external row under this protocol.

\subsection{Ablation Studies in Simulations}
We conduct ablations on the DT split of EVT-Bench using bounding-box prompts, because this setting requires both accurate target grounding and persistent identity maintenance under distractors. Table~\ref{tab:ablation_dt_bbox} isolates four design factors: cross-view fusion (number of views), the action-conditioned latent world-model objective (WM), memory-bank length (Mem.), and waypoint decoder type. The two generative decoders replace our direct regression head with a transformer of the same parameter count under an otherwise identical training recipe, sampling the trajectory by flow matching or by DDIM with two steps in both cases. We do not verify that these choices transfer to the language policy used in the benchmark comparison.

\begin{table}[t]
\centering
\caption{Component ablation on EVT-Bench DT with bounding-box prompts. The default uses the three-view configuration of Section~\ref{sec:setup}; the single-view row is the configuration deployed on the robot.}
\label{tab:ablation_dt_bbox}
\par\vspace{0.35em}
\fontsize{8.5pt}{10pt}\selectfont
\setlength{\tabcolsep}{4pt}
\renewcommand{\arraystretch}{0.95}
\begin{tabular}{@{}lccccccc@{}}
\toprule
Variant & Views & WM & Mem. & Action & SR$\uparrow$ & TR$\uparrow$ & CR$\downarrow$ \\
\midrule
\textbf{Default USS} & 3 & Yes & 16 & Ours & 83.6 & 81.5 & \textbf{2.9} \\
\midrule
Single view & 1 & Yes & 16 & Ours & 78.1 & 76.3 & 3.1 \\
w/o WM & 3 & No & 16 & Ours & 80.4 & 79.2 & 3.0 \\
Mem.\ 0 & 3 & Yes & 0 & Ours & 72.2 & 73.3 & 3.4 \\
Mem.\ 32 & 3 & Yes & 32 & Ours & \textbf{84.1} & \textbf{82.3} & \textbf{2.9} \\
Flow-match & 3 & Yes & 16 & FM & 78.8 & 79.9 & 3.6 \\
DDIM & 3 & Yes & 16 & DDIM & 75.8 & 76.4 & 4.7 \\
\bottomrule
\end{tabular}
\end{table}

Table~\ref{tab:ablation_dt_bbox} quantifies the evaluated design choices. Removing temporal memory causes the largest degradation (11.4 SR and 8.2 TR points), consistent with the role of explicit identity memory under distractors reported for TrackVLA++~\citep{zhang2025trackvlapp}. Cross-view fusion contributes 5.5 SR and 5.2 TR, the second-largest effect in the table; the single-view instance, which is the configuration deployed on the robot, keeps an essentially unchanged collision rate. Removing latent prediction decreases SR and TR by 3.2 and 2.3 points. Extending memory from 16 to 32 frames is the strongest configuration in the table, but it gains only 0.5 SR and 0.8 TR while doubling the number of memory tokens the fusion encoder attends over; we therefore keep 16 frames as the default, as the better accuracy-cost trade-off. Flow matching and DDIM reduce SR by 4.8 and 7.8 points, respectively, and increase collision rate, so direct waypoint prediction outperforms the two evaluated generative decoders~\citep{chi2025diffusion,bjorck2025gr00t} on tracking, though these outcomes alone do not establish a latency advantage for it.


\section{Conclusion}
\label{sec:conclusion}
This paper introduces unified spatial-semantic prompting for embodied visual tracking, one formulation and one architecture in which a description, a point, a box, and a mask are interchangeable specifications supplied once by the person, with one instance trained per interface. Across 320 physical trials the interfaces separate in the single scene where a distractor shares most of the target's described attributes, which is where designating an instance earns its requirement that the target be visible. On EVT-Bench under the standard language protocol, USS obtains the highest success rate among non-MLLM methods at 57 FPS, below the MLLM trackers in success rate and several times above their reported throughput.

\noindent\textbf{Limitations.} Two limitations remain. First, the current model is trained with relatively well-formed spatial prompts; integrating prompt-noise augmentation or lightweight prompt refinement would make the interfaces more robust to imprecise designation. Second, the locomotion policy is still simple, supporting mainly flat-ground walking without low-level obstacle avoidance or terrain-aware control. A stronger low-level controller would extend USS to more complex deployment environments without changing the proposed spatial-semantic tracking formulation.

\section*{Acknowledgment}
We disclose that the illustrative elements of Figure~\ref{fig:motivation} were generated with ChatGPT Image 2.0 from the prompt ``Create a four-panel scientific illustration of robot target tracking based on the robot image uploaded, using language for an out-of-view target and spatial prompts for similar-looking, unfamiliar, and time-critical targets.'', and were subsequently edited manually by the authors. No experimental data, results, or evaluation images were generated or modified using AI.


\bibliographystyle{assets/plainnat}
\bibliography{uss}

\ifwithappendix
\clearpage
\beginappendix
\section{Data Collection}
\label{app:data_collection}

\paragraph{Training data and trajectory labels.}
We construct the simulation training set from the training branch of EVT-Bench introduced by TrackVLA~\citep{wang2025trackvla}. The humanoid assets, URDF models, target identities, and route specifications are inherited from EVT-Bench. We select the 800 episodes with the longest target routes and replay them in Habitat. For each episode, the target humanoid follows the provided start and goal waypoints on the navigation mesh, while distractor humanoids, when present, follow their corresponding benchmark routes. The simulator is rendered with three egocentric cameras at 10 Hz and $512\times512$ resolution.

The waypoint supervision is generated by a teacher tracker in the same simulator. At each frame, the teacher estimates a following point slightly behind the target according to the target position and walking direction, then rolls out a short-horizon path from the current tracker pose toward this point on the navigation mesh. The rollout uses velocity matching with a catch-up term when the tracker falls behind, and applies simple collision-aware steering around other humanoids. The planned future tracker positions are stored in world coordinates and then transformed into the current tracker-centric frame to obtain the ground-truth waypoint sequence used by $\mathcal{L}_{\mathrm{wp}}$. We also record view-wise target visibility labels for the auxiliary presence loss. A view is marked visible if at least one of seven approximate humanoid keypoints is inside the camera frustum and passes an occlusion check. These keypoints cover the head, shoulders, torso, hips, and lower body. For each candidate keypoint, we cast a ray from the camera to the keypoint in Habitat and reject it if scene geometry intersects the ray before the target.

\paragraph{Prompt annotation.}
For language-prompt training, the robot start pose and text instruction are read directly from the original TrackVLA episode annotations. For spatial-prompt training, the initial robot pose is adjusted to ensure the target is visible in the tracker's views at the beginning of the episode, with a heading perturbation to avoid a perfectly centered view. We provide visualizations of the training data for spatial prompts in Appendix~\ref{sec:visualization-training-data}. The point and bounding-box prompts are generated by projecting the target humanoid state into the image. The bounding box is computed by projecting the eight corners of a coarse upright 3D cuboid around the humanoid. The point prompt is obtained by sampling a conservative 3D point around the humanoid torso, relative to the humanoid base position, and projecting that point into the camera image using the same camera intrinsics and sensor pose. This gives stable first-frame spatial prompts without requiring per-frame manual annotation. The mask prompt is obtained by applying SAM2 to the first-frame target region and is used as the first-frame target mask. These first-frame annotations are kept fixed as the target specification throughout the rollout.

\section{Training Details}
\label{app:training_details}

\paragraph{Optimization.}
USS is trained by imitation losses on future waypoints, view-wise target visibility, and the latent dynamics objective described in the main paper. The PE-Spatial-B16-512 image encoder and PE-Core-B16-224 text encoder are initialized from pretrained weights, the text encoder is frozen, and only the last two image-transformer blocks are fine-tuned. We train with AdamW for 12 epochs using a learning rate of $5\times10^{-5}$, weight decay $0.01$, and the default AdamW momentum parameters $\beta_1=0.9$, $\beta_2=0.999$, and $\epsilon=10^{-8}$. Backbone parameters use a $0.1$ learning-rate multiplier. We use a cosine learning-rate schedule, batch size 1 per GPU, and gradient clipping with norm 35. After the supervised training stage, we further collect model-driven rollouts and relabel them with the same teacher planner in a DAgger-style stage. This exposes the policy to states induced by its own actions and reduces the distribution shift of pure imitation learning. One instance is trained per prompt modality under this identical recipe. The main model hyperparameters are summarized in Table~\ref{tab:appendix_training_hparams}.

\begin{table}[t]
\centering
\caption{Main model hyperparameters for USS.}
\label{tab:appendix_training_hparams}
\scriptsize
\par\vspace{0.5em}
\setlength{\tabcolsep}{3pt}
\renewcommand{\arraystretch}{1.05}
\begin{tabular}{@{}p{0.30\textwidth}cp{0.31\textwidth}c@{}}
\toprule
Model hyperparameter & Value & Model hyperparameter & Value \\
\midrule
Input views $N$ (simulation) & 3 & Sparse queries per view $K_q$ & 10 \\
Input views $N$ (real robot) & 1 & Planner hidden dimension $C_v$ & 256 \\
Future waypoint horizon $M$ & 10 & Prompt embedding dimension & 256 \\
PE-Spatial token dimension & 768 & Prompt-fusion layers / heads & 2 / 8 \\
BBox prompt tokens & 9 & BBox ROIAlign size & $7\times7$ \\
Point ROIAlign size & $5\times5$ & Point pseudo-box side length & 160 px \\
Point prompt tokens & 4 & Memory-bank layers / heads & 2 / 8 \\
Memory-bank length & 16 frames & World-model FFN dimension & 1024 \\
World-model layers / heads & 2 / 8 & World-model loss weight $\lambda_{\mathrm{wm}}$ & 0.2 \\
Presence loss weight $\lambda_{\mathrm{pres}}$ & 0.05 & Input resolution & $512\times512$ \\
3D position depth bins & 64 & Fine-tuned image blocks & last 2 \\
\bottomrule
\end{tabular}
\end{table}

\section{Inference Details}
\paragraph{Inference.}
The model predicts a future waypoint sequence in the tracker-centric ground-plane frame and view-wise target-presence logits. We execute only the first predicted waypoint at each control step. This waypoint is transformed from the tracker frame to world coordinates, snapped to the navigation mesh through Habitat, and applied with the same safe tracker-motion routine used by the rollout runner. When all predicted view-wise presence probabilities are below $0.5$, evaluation optionally falls back to the next waypoint from the previous predicted world trajectory, which stabilizes short target disappearances without changing the initial prompt. The EMA representation encoder and the latent world-model predictor are removed at inference time, so the auxiliary branch adds no runtime cost.

\section{Real-World Deployment Details}
\label{app:real_world_deployment}
\begin{figure}[H]
    \centering
    \includegraphics[width=0.75\linewidth]{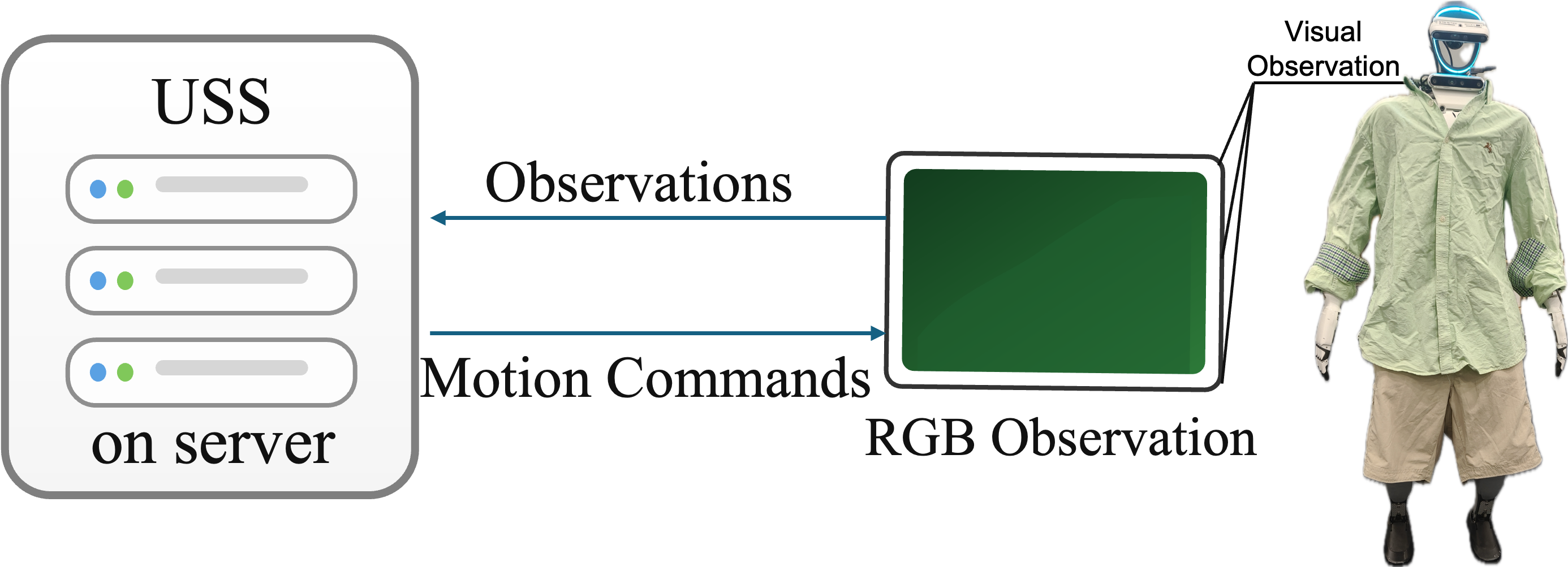}
    \caption{Real-world deployment platform.}
    \label{fig:robot_platform}
\end{figure}
\paragraph{Robot platform.}
We deploy USS on a Unitree G1 humanoid robot for real-world embodied visual tracking. The robot uses an Intel RealSense D455 camera mounted on its chest to capture egocentric RGB observations. Because the platform carries a single camera, all real-robot results use a single-view instance of the same architecture, training recipe, and supervision as the simulation policies. Given the current visual observation and a target prompt, USS predicts the motion commands required for target following, enabling the robot to continuously track and follow the specified target in real-world indoor environments.

\paragraph{Real-world system deployment.}
In our real-world experiments, USS runs on a remote server equipped with an NVIDIA RTX 4090 graphics processing unit (GPU). The Unitree G1 robot communicates with the server through Ethernet. At each control step, the robot first captures the current egocentric RGB image using the chest-mounted RealSense D455 camera and transmits the visual observation to the server. The server then receives the image and the target prompt, performs USS inference, and generates the motion command for target following. The predicted command is transmitted back to the robot through Ethernet and executed by the low-level controller, forming a closed-loop target-following system.

\paragraph{Control and execution.}
USS is responsible for high-level target perception and following decision-making, rather than joint-level control or low-level gait generation. Specifically, the server-side model predicts a robot-frame waypoint trajectory, which is converted into a velocity command by using the first waypoint displacement over a fixed time horizon for linear velocities $(v_x, v_y)$ and the heading direction to the final waypoint over the same horizon for yaw rate $v_\omega$. The resulting command $(v_x, v_y, v_\omega)$ is sent to the robot-side low-level controller for gait generation and execution. This design decouples target specification, visual perception, and high-level following decisions from low-level motion execution, allowing our method to focus on target identity preservation and closed-loop tracking under different prompt types. The latent dynamics branch used during training is removed during deployment, so it introduces no additional computational overhead during real-world inference.

\paragraph{Real-world scenes.}
We evaluate USS in four indoor real-world scenes: a narrow corridor, a pedestrian-distractor scene, a long-route scene, and a similar-person scene. These scenes cover different levels of difficulty, ranging from basic closed-loop following to challenging target identity preservation. The narrow-corridor scene evaluates basic following ability in a constrained space. The pedestrian-distractor scene introduces unrelated pedestrians and tests whether the model can maintain the intended target identity in a dynamic environment. The long-route scene includes a longer trajectory with turns, testing temporal stability and accumulated control errors during real-world closed-loop execution. The similar-person scene contains two people of similar appearance: both wear a black top and only the trousers differ, so the language description still identifies the target uniquely while providing little distinguishing evidence, which isolates how reliably a sufficient description is grounded and held.

\paragraph{Prompt protocol.}
In the real-world experiments, we evaluate four types of target prompts: language, bounding-box, point, and mask prompts. For language prompts, the target is specified by a natural-language description. For the three spatial prompts, the target is specified on the first-frame image: a bounding box drawn around the target, a single point on the target, or a segmentation mask of the target. The prompt is provided only once at the beginning of each trial. After initialization, no additional human correction, target re-specification, or prompt update is provided. In every condition the target is visible and unoccluded at initialization, matching the spatial-prompt initialization used in simulation, so the conditions differ only in how the instance is designated. The robot must then rely on its continuous visual observations, server-side USS inference, and low-level controller execution to follow the initially specified target. All real-robot trials use the simulation-trained policies zero-shot, with no real-world data, fine-tuning, or calibration.

\paragraph{Evaluation protocol.}
For each scene and each prompt type, we conduct 20 real-world trials, 320 in total. A trial fails if the robot switches to another person, or turns away and then either loses the target for more than five seconds or never recovers the heading needed to resume; all remaining trials count as successes. We report the success rate for each scene and each prompt type to compare the real-world tracking performance of language and spatial (bounding-box, point, and mask) target specification.

\paragraph{Practical observations.}
We observe that spatial prompts are particularly effective when multiple similar target candidates are present. In such scenarios, a language description may correctly describe several people but still fail to uniquely identify the intended physical instance, and even a description that does identify the target uniquely may provide little distinguishing evidence. In contrast, a spatial prompt provides explicit instance-level visual evidence at the beginning of the trial, helping the robot preserve the intended target identity during subsequent closed-loop tracking. This effect is especially clear in the similar-person scene, where language prompts are more likely to cause target confusion.

\clearpage
\section{Visualization Of Training Data}
\label{sec:visualization-training-data}
\setlength{\intextsep}{4pt plus 1pt minus 1pt}

\begin{figure}[H]
    \centering
    \begin{minipage}[t]{0.48\textwidth}
        \centering
        \includegraphics[width=0.32\linewidth]{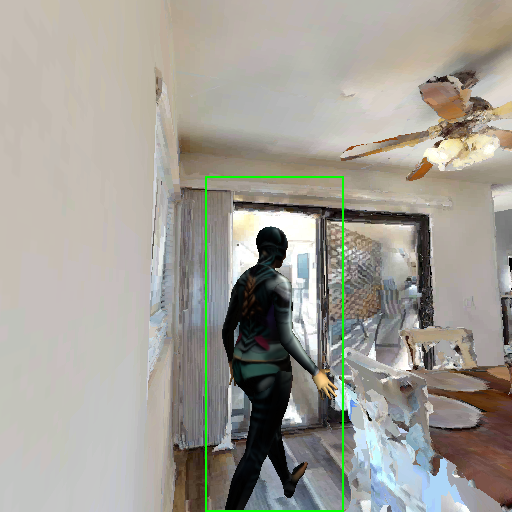}
        \includegraphics[width=0.32\linewidth]{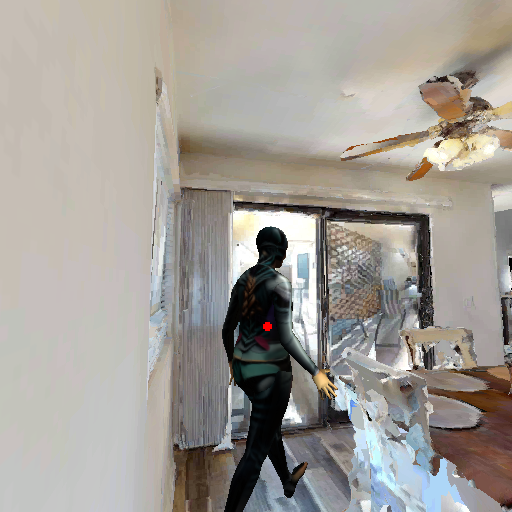}
        \includegraphics[width=0.32\linewidth]{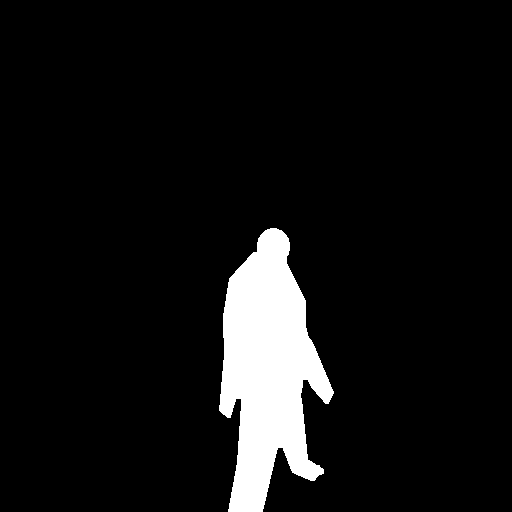}\\[-0.2em]
        {\small (a) Spatial prompts 1}
    \end{minipage}
    \hfill
    \begin{minipage}[t]{0.48\textwidth}
        \centering
        \includegraphics[width=0.32\linewidth]{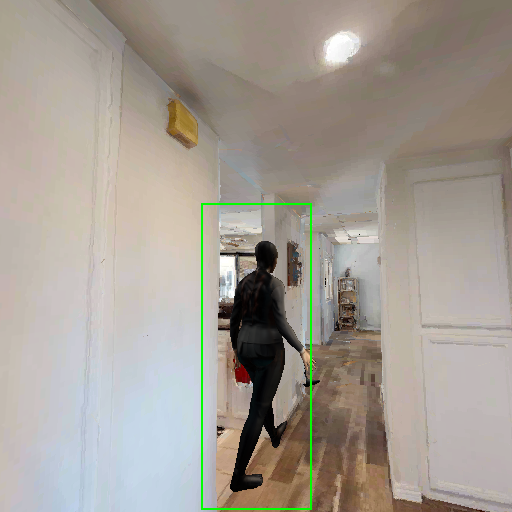}
        \includegraphics[width=0.32\linewidth]{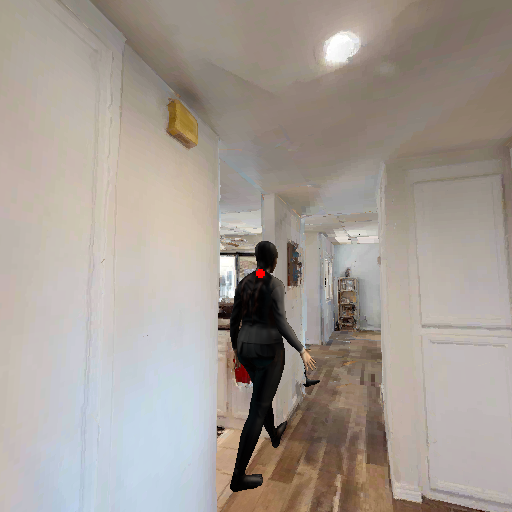}
        \includegraphics[width=0.32\linewidth]{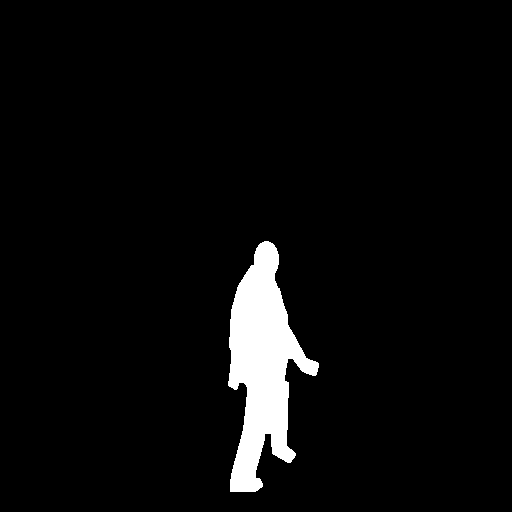}\\[-0.2em]
        {\small (b) Spatial prompts 2}
    \end{minipage}
    \caption{Examples of spatial prompts used in the training data. Each group shows bounding-box, point, and mask annotations from left to right.}
    \label{fig:training-data-spatial-prompts}
\end{figure}

\begin{figure}[H]
\centering
\includegraphics[width=0.19\linewidth]{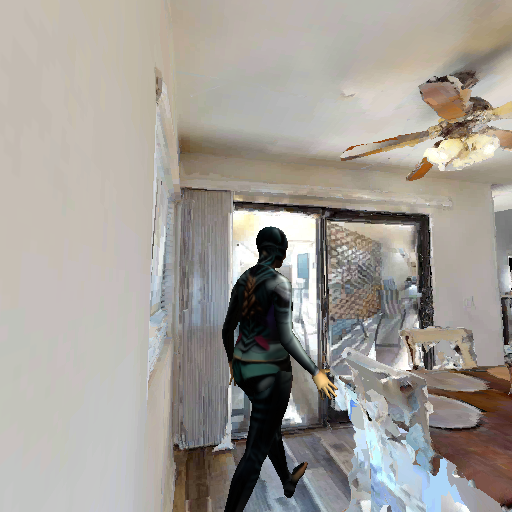}
\includegraphics[width=0.19\linewidth]{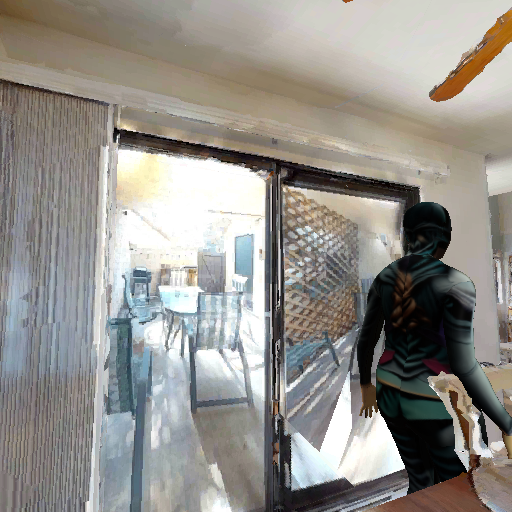}
\includegraphics[width=0.19\linewidth]{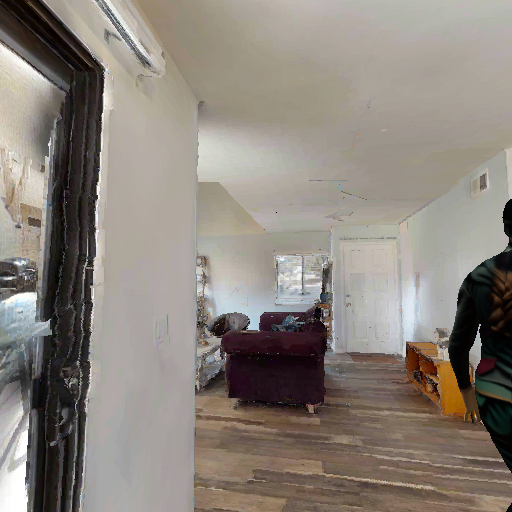}
\includegraphics[width=0.19\linewidth]{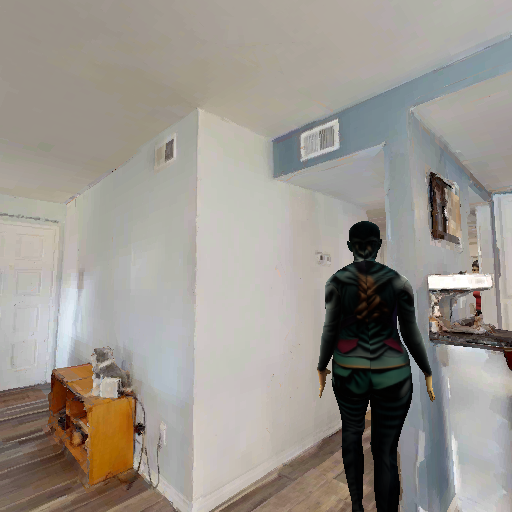}
\includegraphics[width=0.19\linewidth]{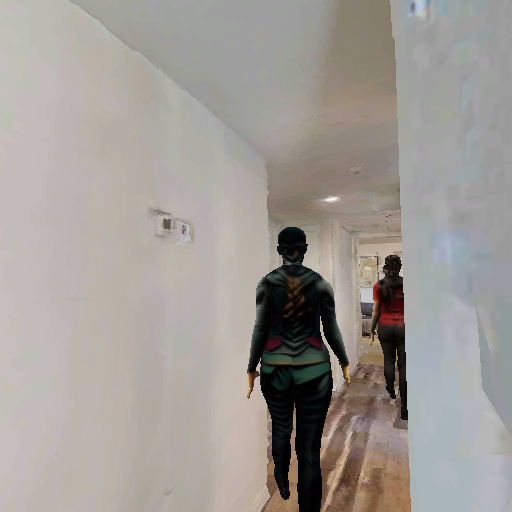}\\[-0.2em]
{\small (a) Front-view observation 1.}

\vspace{0.3em}
\includegraphics[width=0.19\linewidth]{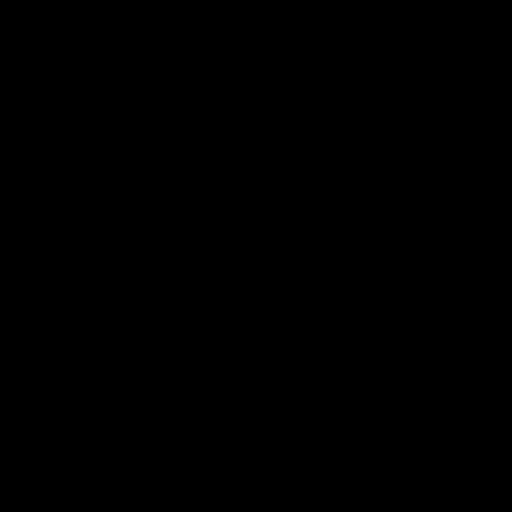}
\includegraphics[width=0.19\linewidth]{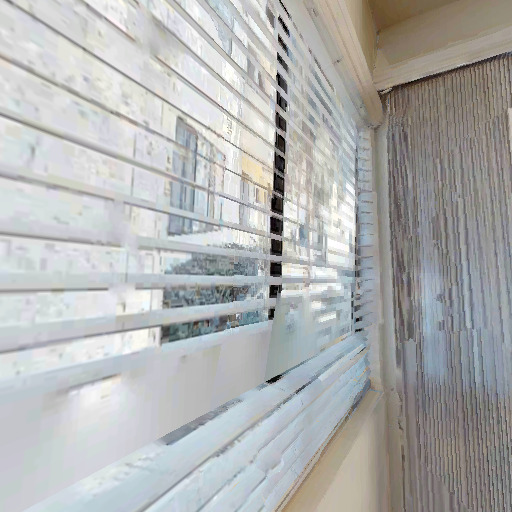}
\includegraphics[width=0.19\linewidth]{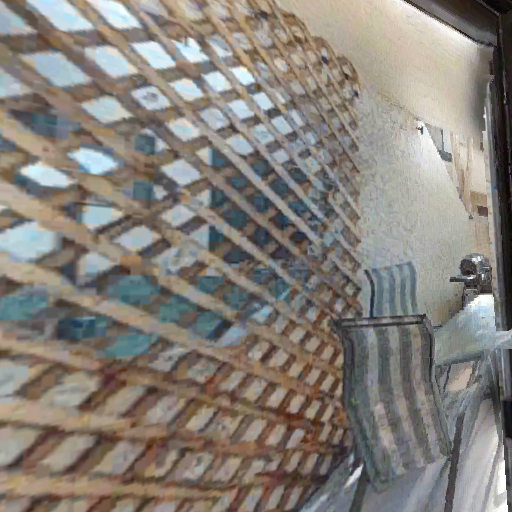}
\includegraphics[width=0.19\linewidth]{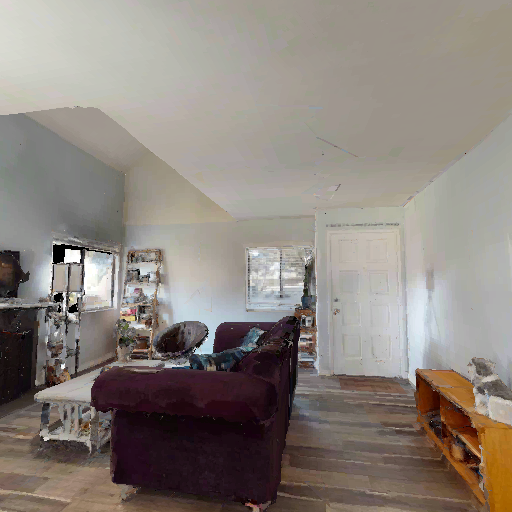}
\includegraphics[width=0.19\linewidth]{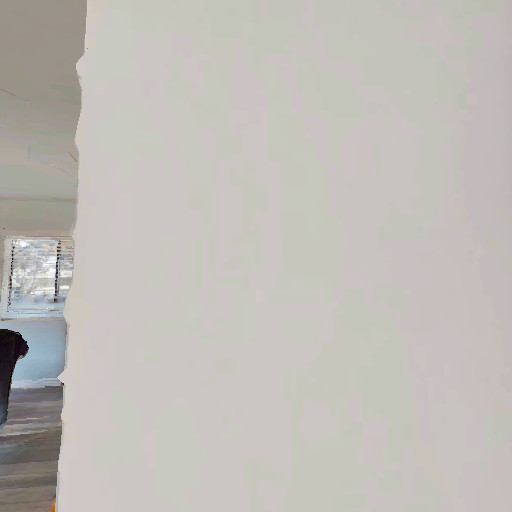}\\[-0.2em]
{\small (b) Left-view observation 1.}

\vspace{0.3em}
\includegraphics[width=0.19\linewidth]{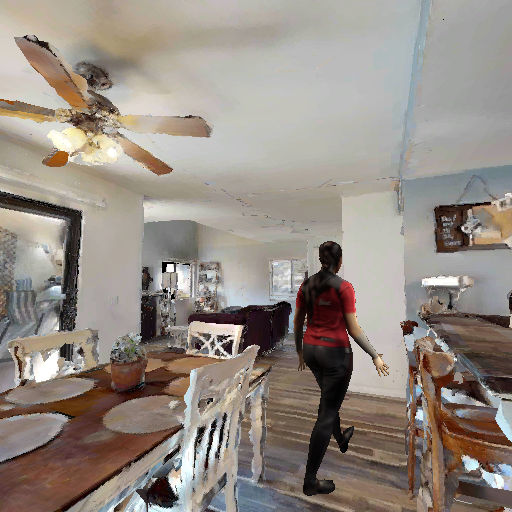}
\includegraphics[width=0.19\linewidth]{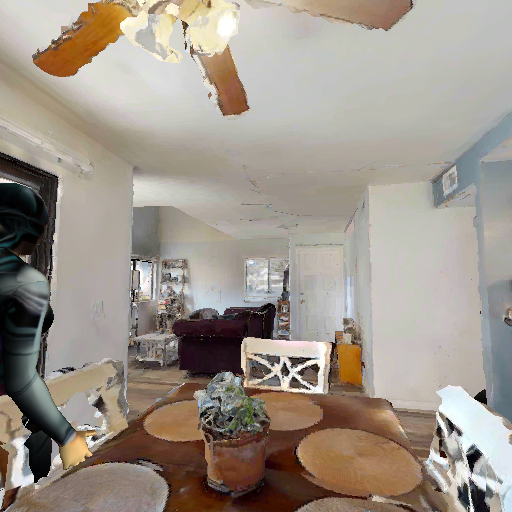}
\includegraphics[width=0.19\linewidth]{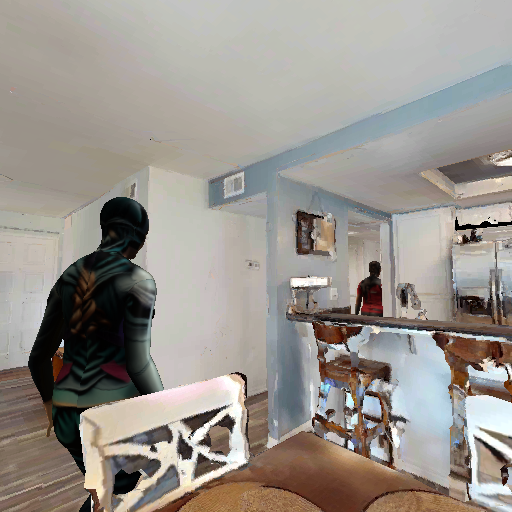}
\includegraphics[width=0.19\linewidth]{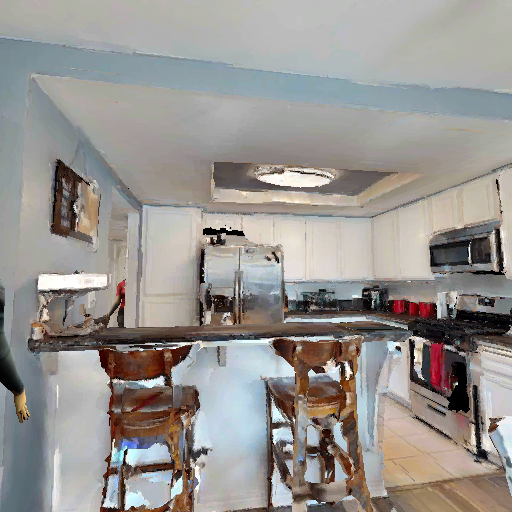}
\includegraphics[width=0.19\linewidth]{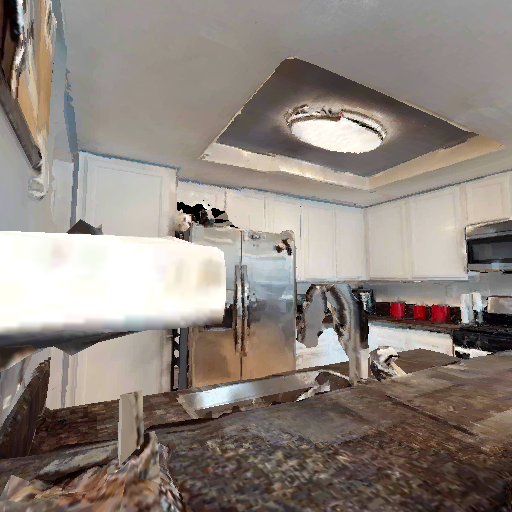}\\[-0.2em]
{\small (c) Right-view observation 1.}

\caption{Temporal observations from the tracker (training episode 1).}
\label{fig:training-data-ep8-observations}
\end{figure}

\begin{figure}[H]
\centering
\includegraphics[width=0.19\linewidth]{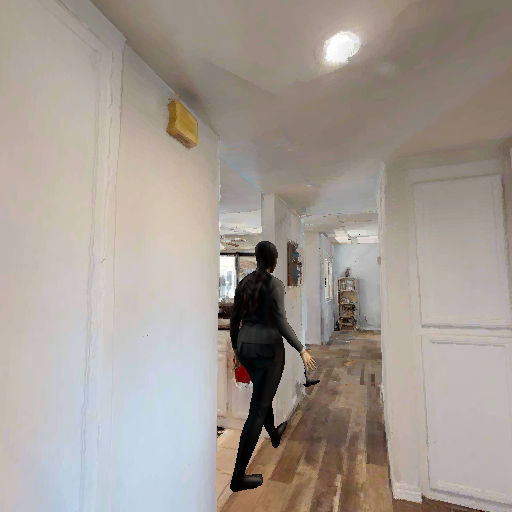}
\includegraphics[width=0.19\linewidth]{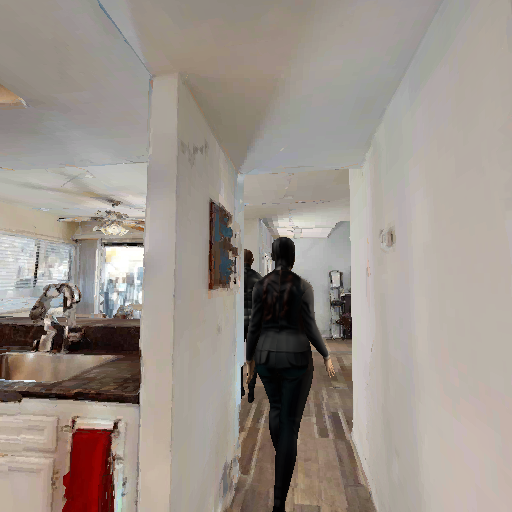}
\includegraphics[width=0.19\linewidth]{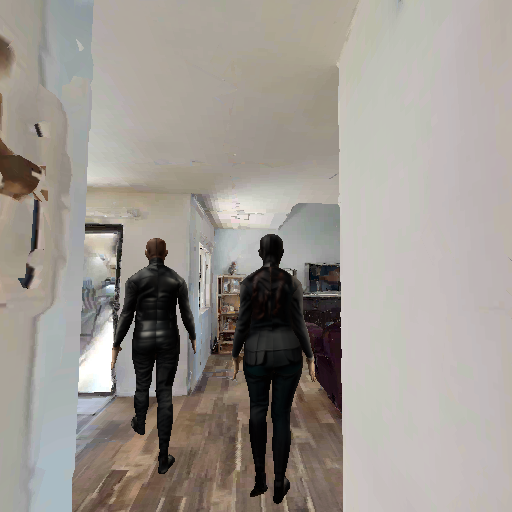}
\includegraphics[width=0.19\linewidth]{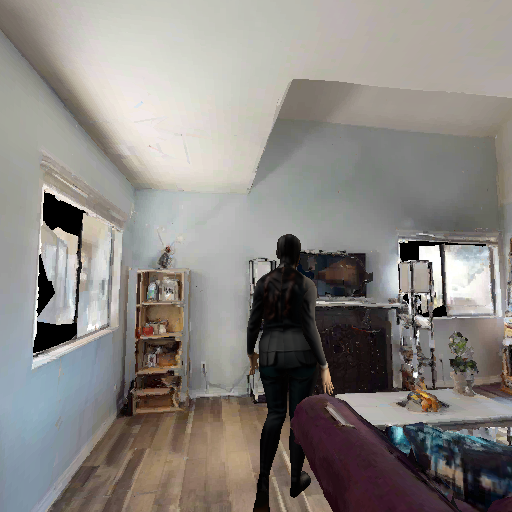}
\includegraphics[width=0.19\linewidth]{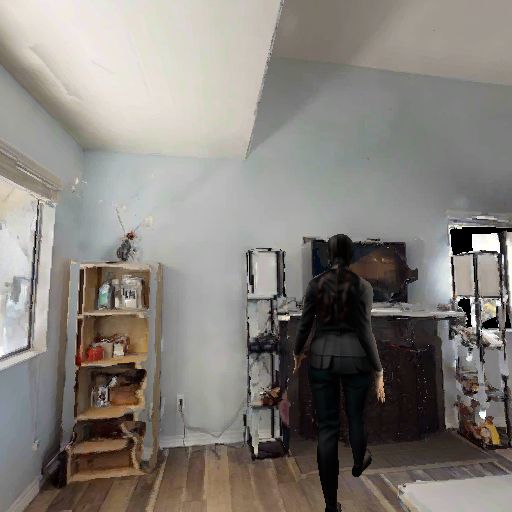}\\[-0.2em]
{\small (a) Front-view observation 2.}

\vspace{0.3em}
\includegraphics[width=0.19\linewidth]{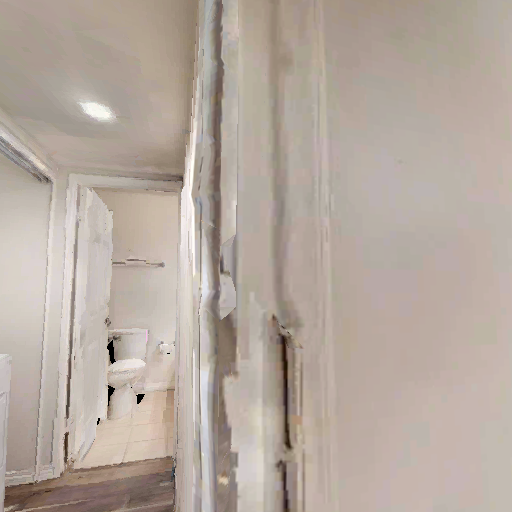}
\includegraphics[width=0.19\linewidth]{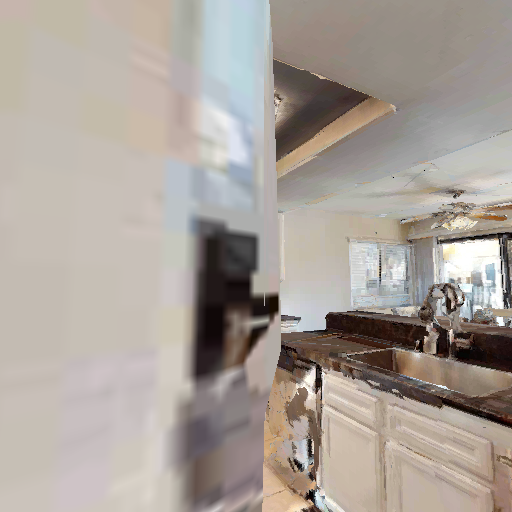}
\includegraphics[width=0.19\linewidth]{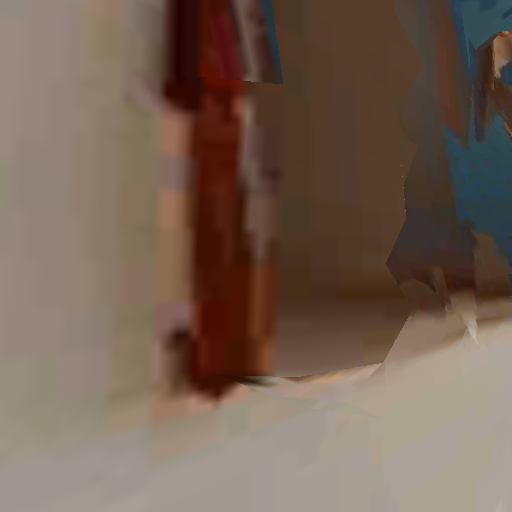}
\includegraphics[width=0.19\linewidth]{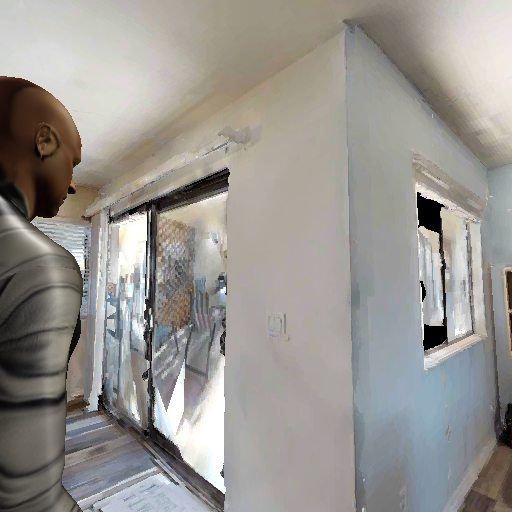}
\includegraphics[width=0.19\linewidth]{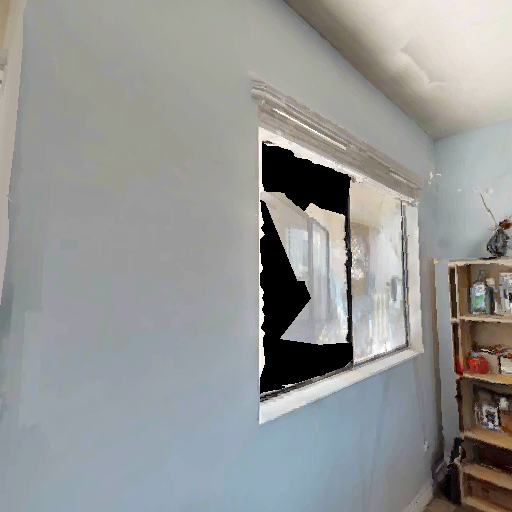}\\[-0.2em]
{\small (b) Left-view observation 2.}

\vspace{0.3em}
\includegraphics[width=0.19\linewidth]{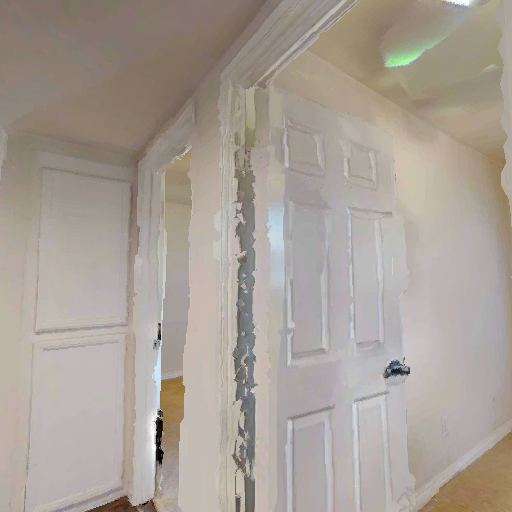}
\includegraphics[width=0.19\linewidth]{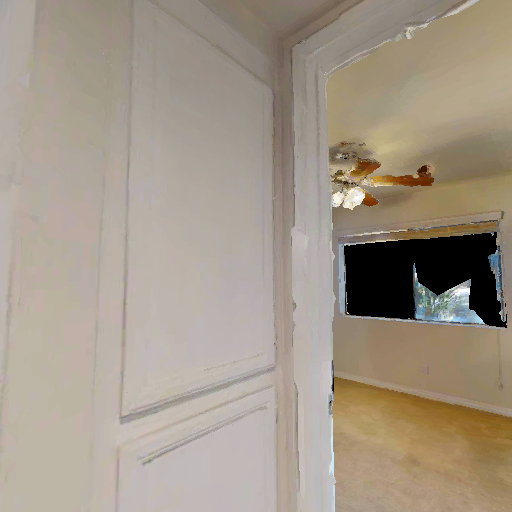}
\includegraphics[width=0.19\linewidth]{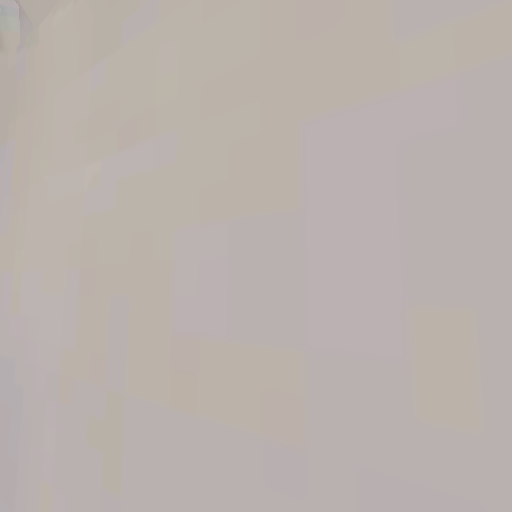}
\includegraphics[width=0.19\linewidth]{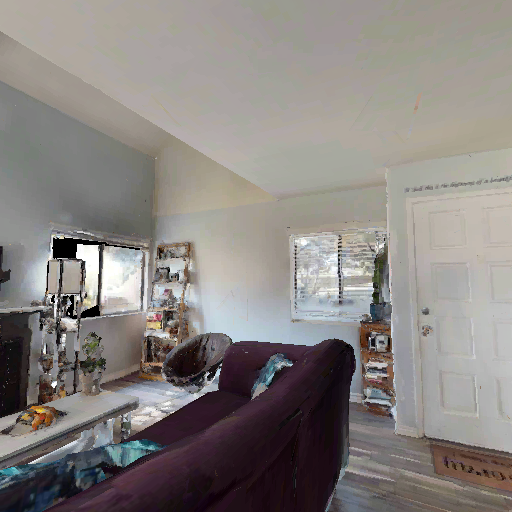}
\includegraphics[width=0.19\linewidth]{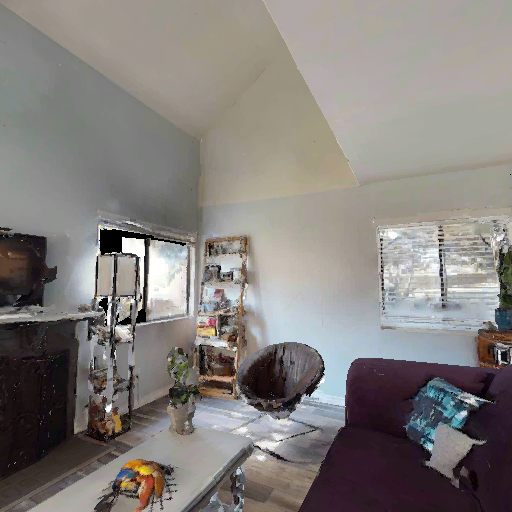}\\[-0.2em]
{\small (c) Right-view observation 2.}

\caption{Temporal observations from the tracker (training episode 2).}
\label{fig:training-data-ep3-observations}
\end{figure}

\fi

\end{document}